\documentclass[lettersize,journal]{IEEEtran}
\usepackage{amsmath,amsfonts,bm,bbm}
\usepackage{amsthm}
\usepackage{algcompatible}
\usepackage{algorithm}
\usepackage{array}
\usepackage{textcomp}
\usepackage{booktabs}
\usepackage{stfloats}
\usepackage{url}
\usepackage{verbatim}
\usepackage{multirow}
\usepackage{makecell}

\usepackage{graphicx}
\usepackage{soul, color,xcolor}

\usepackage{cite}

\usepackage{orcidlink}
\usepackage{algpseudocode}

\usepackage{multirow}
\usepackage{colortbl}

\usepackage{pifont}

\usepackage{upgreek}
    
\begin{document}

\title{Learning Multi-view Anomaly Detection with Efficient Adaptive Selection}

\author{Haoyang He, Jiangning Zhang, Guanzhong Tian, Chengjie Wang, Lei Xie
\IEEEcompsocitemizethanks{
\IEEEcompsocthanksitem Haoyang He, Jiangning Zhang and Lei Xie are with the State Key Laboratory of Industrial Control Technology, Zhejiang University, Hangzhou 310027, China (e-mail: haoyanghe@zju.edu.cn; 186368@zju.edu.cn; leix@iipc.zju.edu.cn). 
\IEEEcompsocthanksitem Guanzhong Tian is with the Ningbo Innovation Center, Zhejiang University, Hangzhou 310027, China (e-mail: gztian@zju.edu.cn).
\IEEEcompsocthanksitem Chengjie Wang is with the YouTu Lab, Tencent, Shanghai 200233, China (e-mail: jasoncjwang@tencent.com).
}
\thanks{Corresponding authors: Jiangning Zhang and Lei Xie}
}

\markboth{IEEE TRANSACTIONS ON Multimedia}%
{Shell \MakeLowercase{\textit{et al.}}: A Sample Article Using IEEEtran.cls for IEEE Journals}


\maketitle

\begin{abstract}
This study explores the recently proposed and challenging multi-view Anomaly Detection (AD) task. Single-view tasks will encounter blind spots from other perspectives, resulting in inaccuracies in sample-level prediction. Therefore, we introduce the \textbf{M}ulti-\textbf{V}iew \textbf{A}nomaly \textbf{D}etection (\textbf{MVAD}) approach, which learns and integrates features from multi-views. Specifically, we propose a \textbf{M}ulti-\textbf{V}iew \textbf{A}daptive \textbf{S}election (\textbf{MVAS}) algorithm for feature learning and fusion across multiple views. The feature maps are divided into neighbourhood attention windows to calculate a semantic correlation matrix between single-view windows and all other views, which is an attention mechanism conducted for each single-view window and the top-$k$ most correlated multi-view windows. Adjusting the window sizes and top-$k$ can minimise the complexity to $O((hw)^\frac{4}{3})$. Extensive experiments on the Real-IAD dataset under the multi-class setting validate the effectiveness of our approach, achieving state-of-the-art performance with an average improvement of \textbf{+2.5$\uparrow$} across \textbf{10 metrics} at the sample/image/pixel levels, using only \textbf{18M} parameters and requiring fewer FLOPs and training time. The codes are available at \url{https://github.com/lewandofskee/MVAD}.
\end{abstract}


\section{Introduction}
\IEEEPARstart{A}{nomaly} detection (AD) is a critical application within computer vision, focusing on identifying anomalies to ensure quality and mitigate potential risks~\cite{survey3}. This task is widely applicable in industrial~\cite{survey1,survey2}, medical~\cite{medical1}, and video surveillance~\cite{wu2023dss, chang2021contrastive, xu2019video,chu2018sparse,song2019learning} AD. Diverse AD datasets have been curated to cater to various scenarios, encompassing 2D~\cite{mvtec,visa}, 2D with depth maps~\cite{mvtec3d}, and 3D datasets~\cite{real3d}. Recently, the Real-IAD~\cite{realiad} dataset was introduced for multi-view AD. In contrast to traditional single-view 2D images or 3D point cloud data, the multi-view images in this dataset offer multiple perspectives of each object, where anomalies may manifest in one view while appearing normal in others due to interrelations among different views. This work addresses the intricate task of visual industrial multi-view anomaly detection.

\begin{figure*}[t]
\centering
\includegraphics[width=1\textwidth]{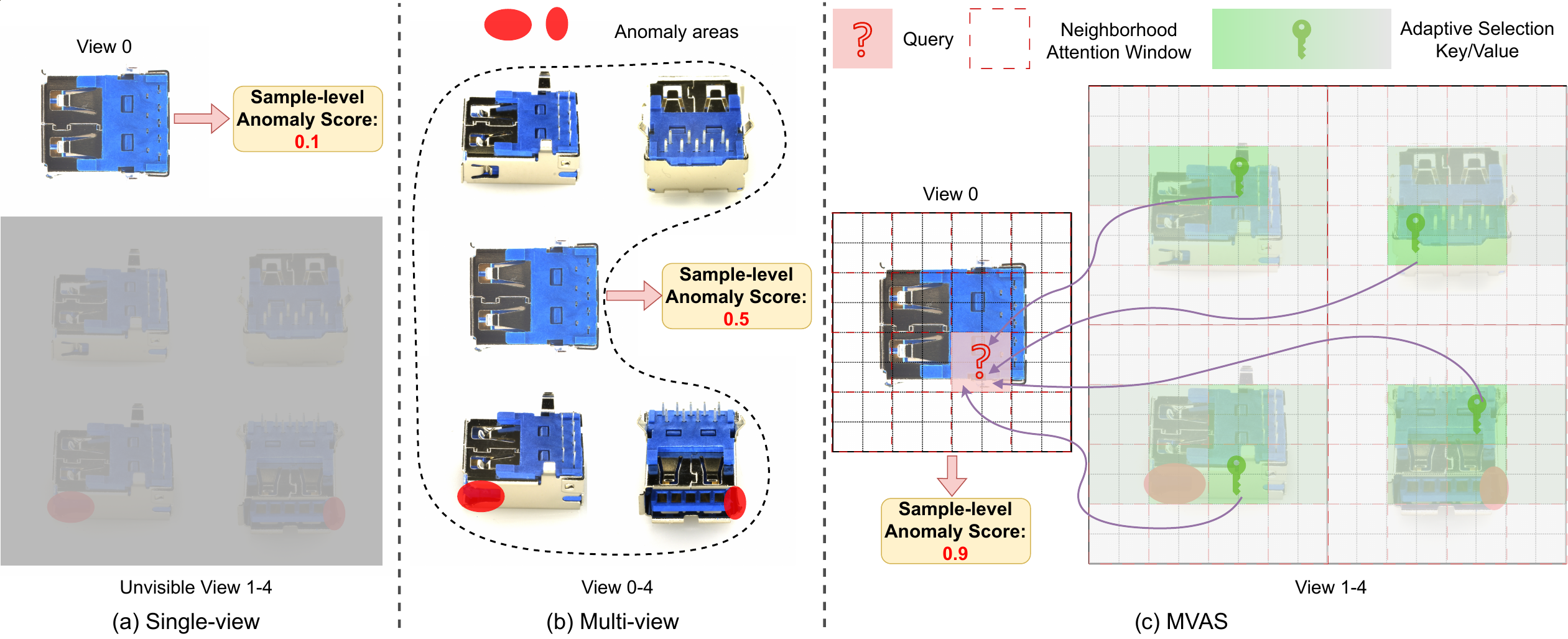} 
\vspace{-0.5em}
\caption{ \textbf{\textit{a)}} Single-view of a sample input. \textbf{\textit{b)}} Multi-view of a sample input. \textbf{\textit{c)}} The proposed MVAS for multi-view learning and fusing.}
\label{motivation}
\end{figure*}

In traditional single-view tasks, as depicted in Fig. \ref{motivation}-(a), the current single-view is isolated from other views, leading to predictions of normal in the current view despite the actual anomaly present in the sample. Therefore, the concept of multi-view AD is proposed, as illustrated in Fig. \ref{motivation}-(b) with a detailed definition in Sec. \ref{defination}. Anomaly scores are computed for each view, and the maximum score across all views is selected as the final anomaly score for this sample. Real-IAD~\cite{realiad} endeavours to employ existing AD methods to address the task of multi-view AD. Although ~\cite{realiad} constructs a multi-view AD dataset, it only conducts experiments with existing methods in the multi-view setting of the dataset, without proposing algorithm designs specifically for multi-view AD tasks.
Current 2D AD methods can be broadly classified into three categories. \textit{1)} Data augmentation-based methods~\cite{cutpaste,draem,destseg,simplenet}. \textit{2)} Reconstruction-based methods~\cite{ocrgan,huang2022self,ye2020attribute,rd4ad,diad}. \textit{3)} Embedding-based methods~\cite{patchcore,cfa,cflow,pyramidflow}.
Previous research on multi-view AD~\cite{mvadr1,mvadr2} proposed methods for text-based scenarios like web pages and movie reviews, focusing only on anomaly classification without localization. \cite{multiviewad} introduced four solutions for image-based multi-view AD. However, their dataset included different modalities of the same RGB image, such as depth maps. Thus, their definition of multi-view images differs from the visual industrial multi-view AD context addressed in this paper.

\begin{figure}[t]
\centering
\includegraphics[width=1\linewidth]{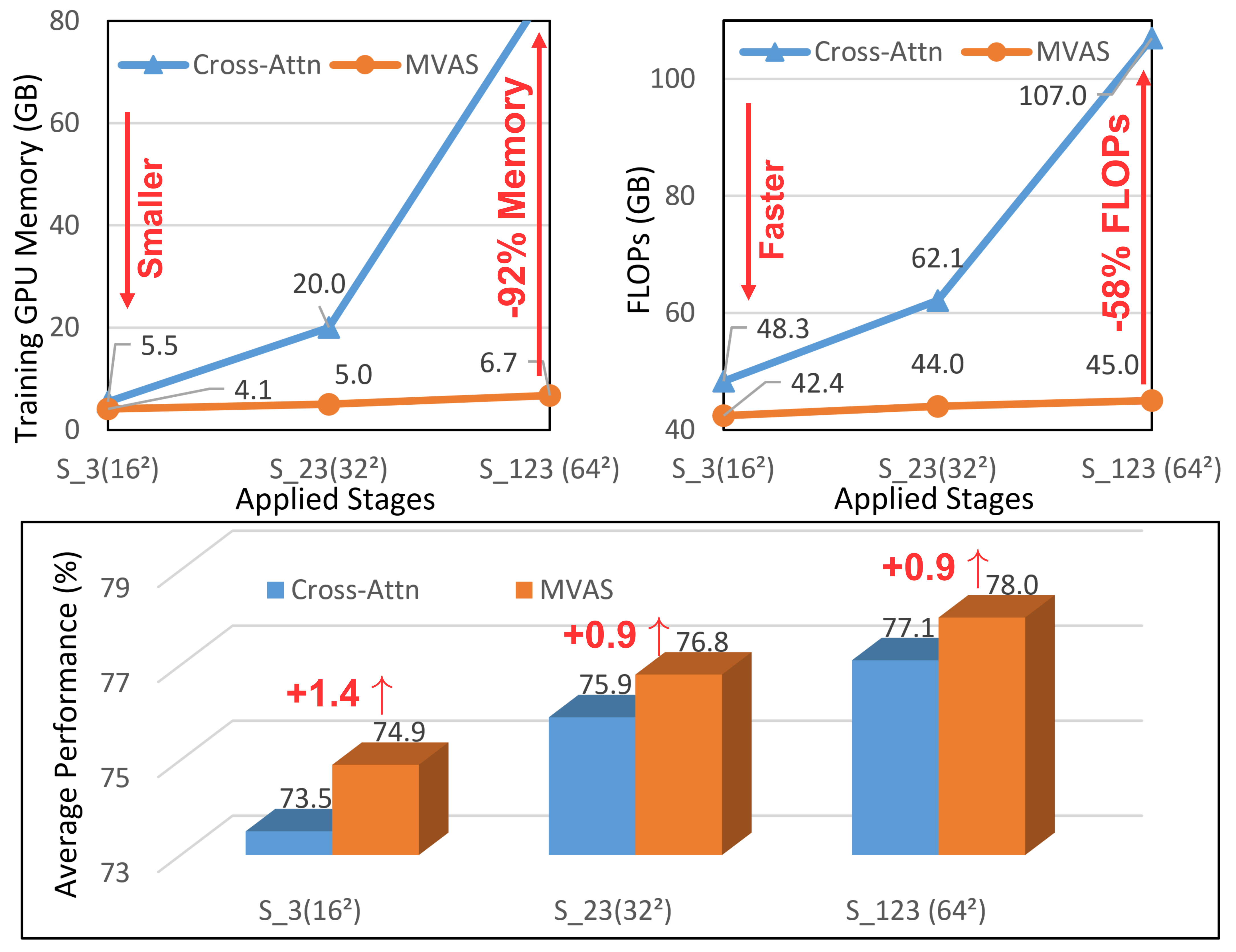}
\vspace{-0.5em}
\caption{A quantitative comparison of cross-attention and the MVAS algorithm at different resolution stages is conducted in terms of \textit{Training memory, FLOPs, and Average Performance} across 10 metrics at the sample/image/pixel levels. As the size of the feature map increases, the FLOPs and training memory required for cross-attention grow with quadratic complexity. In contrast, our MVAS not only exhibits $O((hw)^\frac{4}{3})$ computational complexity but also consistently outperforms cross-attention across different stages.}
\label{crossattn}
\end{figure}

There have been numerous multi-view fusion algorithms in fields such as hand pose estimation~\cite{hand}, 3D scene reconstruction~\cite{3drec}, and autonomous driving~\cite{bevformer}. However, most of these scenarios have explicit camera angle parameters that facilitate effective alignment of the same target area across different views. However, the scenario addressed in this paper lacks camera parameters for different views, making the alignment and fusion more challenging.
Among these multi-view fusion methods~\cite{bevformer,zhou2022cross,ppt,aide,xu2023dmv3d}, the cross-attention mechanism, a simple and effective approach, is widely used for multi-view fusion and learning. Therefore, cross-attention is initially employed to tackle the multi-view AD task. In this approach, one view serves as the query, while the other views act as the key and value for multi-view feature enhancement. However, as illustrated in Fig.~\ref{crossattn}, cross-attention significantly increases FLOPs due to its quadratic computational complexity as the feature map size increases. Additionally, performing attention calculations on all other views' features leads to substantial computational redundancy, resulting in suboptimal outcomes. Consequently, the commonly used cross-attention for multi-view feature interaction faces significant challenges in efficiency and effectiveness when applied to multi-view AD.

To address this limitation and learn the correlations between different views as well as feature fusion across views, we propose the Multi-View Adaptive Selection (MVAS) attention algorithm in Sec.~\ref{mvas}, as shown in Fig. \ref{motivation}-(c). The proposed MVAS divides the input image features into neighbourhood attention windows. Then, the multi-view windows adaptive selection algorithm is implemented to compute the semantic correlation matrix between each single-view window and the concatenated multi-view windows. Multi-view neighbourhood windows with darker colours indicate more substantial semantic relevance to the corresponding single-view window. The top-$k$ number of window indexes is obtained with the correlation matrix, which equals four windows, as shown in this figure. The top-$k$ most correlated multi-view windows are selected as keys and values, enabling neighbourhood correlative cross-attention between the single-view window query and the corresponding keys and values. By limiting attention computation to only the most correlated windows, the computational complexity is significantly reduced, which is a minimum $O((hw)^\frac{4}{3})$ complexity in Sec.~\ref{flopeq}, by altering the window size and number of top-$k$. In Sec.~\ref{ablation}, we validate the theoretical effectiveness and find that setting the top-$k$ value equal to the number of other views yields the best results. Additionally, theoretical calculations of the window size that achieves the minimum FLOPs further confirm the method's effectiveness. Extensive quantitative and qualitative experiments validate the reliability and effectiveness of the approach.
Fig.~\ref{crossattn} presents a comparison between our MVAS and the commonly used cross-attention multi-view fusion algorithm. MVAS demonstrates significant improvements in both training memory and FLOPs efficiency compared to cross-attention. Additionally, it shows substantial enhancements in average performance metrics. Overall, MVAS achieves a good trade-off between effectiveness and efficiency.
Based on the MVAS algorithm, we propose a multi-view anomaly detection (MVAD) approach, as illustrated in Fig. \ref{mvad}. This network comprises a pre-trained encoder for extracting features at different scales. Subsequently, three MVAS blocks of varying scales act on these encoder features to facilitate multi-view feature fusion, resulting in enhanced features for each view. The strengthened multi-view features then pass through an FPN-like network, where information of different scales is fused through convolutional downsampling. The fused features are learned and restored by a decoder with a structure and dimension number equivalent to the encoder. Finally, the features corresponding to different scales from the encoder and decoder are used to calculate MSE losses, which are summed to form the ultimate training loss. Our contributions are summarized as follows:

- We propose a novel approach, MVAD, for multi-view anomaly detection, which firstly addresses multi-view learning in anomaly detection.

- We introduce the MVAS algorithm, which adaptively selects the most semantically correlated neighbouring windows in multi-view for each window in a single-view through attention operations, enhancing detection performance with minimum $O((hw)^\frac{4}{3})$ computational complexity.

- Through theoretical analysis of the lower and upper bounds of FLOPs and extensive experiments, we demonstrate the efficiency of the proposed MVAS algorithm.

- We conducted experiments on the multi-view Real-IAD dataset in multi-class settings. Abundant experiments demonstrate the superiority of MVAD over SoTA methods on an average improvement of +2.5$\uparrow$ across 10 metrics at the sample/image/pixel levels.

\section{Related Work}
\subsection{Anomaly Detection}
Recently, AD has included the following mainstream settings: zero-/few-shot~\cite{winclip,clipad,gpt-4v-ad,patchcore}, noisy learning~\cite{noisy3,noisy5}, and multi-class AD~\cite{uniad,diad,vitad,mambaad,ader}. The unsupervised anomaly detection method mainly includes three methodologies:

\textbf{\textit{1)}} Data augmentation-based methods have shown the potential to enhance the precision of anomaly localization by incorporating synthetic anomalies during the training phase. 
DRAEM~\cite{draem} generates anomaly samples by utilizing Perlin noise. DeSTSeg~\cite{destseg} adopts a comparable approach to DRAEM for synthesizing anomaly samples but introduces a multi-level fusion of student-teacher networks to reinforce the constraints on anomaly data. Additionally, SimpleNet~\cite{simplenet} generates anomaly features by introducing basic Gaussian noise to normal samples. Despite these advancements, the inability to anticipate and replicate all potential anomaly types and categories prevalent in real-world scenarios limits comprehensive anomaly synthesis.

\textbf{\textit{2)}} Reconstruction-based methods learn the distribution of all normal samples during training and reconstruct anomaly regions as normal during testing. OCR-GAN~\cite{ocrgan} decouples image features into various frequencies and employs a GAN network as a reconstruction model. The remarkable generative capacities demonstrated by recent diffusion models have prompted researchers to employ these models in anomaly detection tasks. DiffAD~\cite{diffad} employs synthetic anomalies with the diffusion model as a reconstruction model alongside an additional discriminative model. DiAD~\cite{diad} introduces a semantically guided network to ensure semantic consistency between the reconstructed and input images. 
Nonetheless, reconstruction-based approaches encounter challenges in effectively reconstructing extensive anomaly areas and demonstrating precision in anomaly localization.

\textbf{\textit{3)}} Embedding-based methods can be further classified into three categories: memory bank~\cite{padim,patchcore}, knowledge distillation~\cite{rd4ad,rd++}, and normalizing flow~\cite{pyramidflow}. PatchCore~\cite{patchcore} constructs a memory bank by approximating a set of features that describe normal sample characteristics through the collection of a coreset. 
During testing, anomaly scores are calculated using nearest neighbour search.
RD4AD~\cite{rd4ad} proposes a teacher-student model based on the reverse knowledge distillation paradigm, effectively addressing the issue of non-distinguishing filters in traditional knowledge distillation frameworks.

\begin{figure*}[htp]
\centering
\includegraphics[width=1\textwidth]{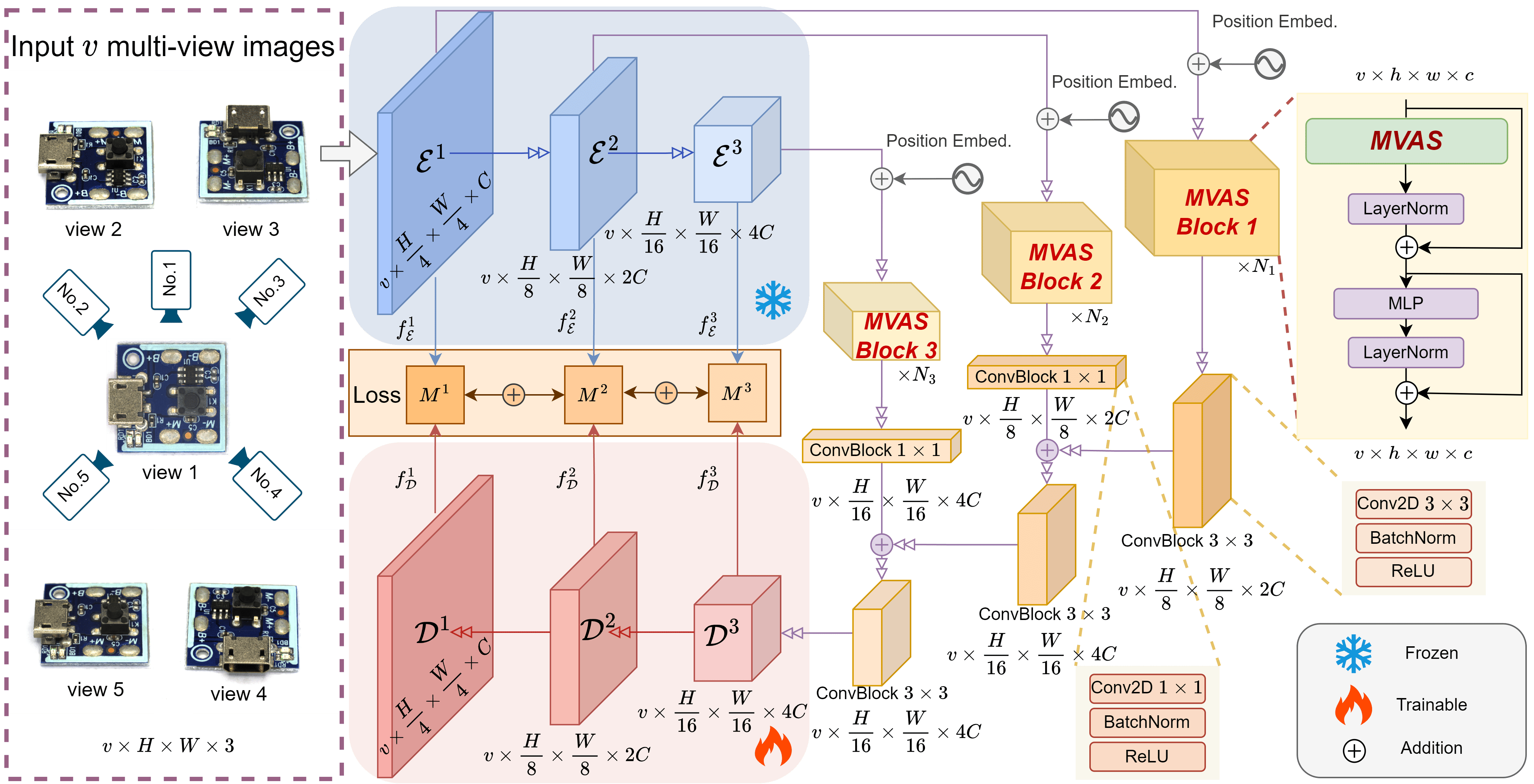} 
\caption{\textbf{The proposed MVAD} comprises a pre-trained teacher encoder $\mathcal{E}$ for multi-scale feature extraction, three distinct dimensional \textit{MVAS Blocks} to enhance single-view features using multi-view data, an FPN-like structure for multi-scale feature aggregation, and a student decoder $\mathcal{D}$ mirroring the encoder's architecture. The training loss is the sum of MSE losses between features extracted at corresponding stages of the encoder $f_{\mathcal{E}}^j$ and decoder $f_{\mathcal{D}}^j$.}
\label{mvad}
\end{figure*}
\subsection{Multi-view Learning}
There have been numerous multi-view fusion algorithms in fields such as hand pose estimation~\cite{hand}, 3D scene reconstruction~\cite{3drec,3drec2,3drec3}, and autonomous driving~\cite{bevformer}. However, most of these scenarios have explicit camera angle parameters that facilitate effective alignment of the same target area across different views.

MVCNN~\cite{mvcnn} introduces a novel CNN-based method for efficiently compressing multi-view information.
MV3D~\cite{mv3d} integrates region-wise features from each view through deep fusion.
ZoomNet~\cite{zoomnet} combines features from different scale views and integrates them through a sequence of convolution operations.
CAVER~\cite{caver} merges RGB and depth view features using cross-attention.
MVDREAM~\cite{mvdream} proposes 3D Self-Attention for fusing multi-view feature information.
AIDE~\cite{aide} introduces two multi-feature fusion methods: cross-attention and adaptive fusion modules.
PPT~\cite{ppt} first extracts feature tokens from each view, concatenates them, and then utilises Self-Attention for feature fusion.
MVSalNet~\cite{mvsalnet} employs element-wise multiplication and addition to merge multi-view features.
MVSTER~\cite{mvster} suggests Epipolar Transformer guided aggregation to effectively capture 2D semantic and 3D spatial correlations.
FLEX~\cite{flex} integrates multi-view convolution layers to capture features from multiple perspectives.
PatchmatchNet~\cite{patchmatchnet} utilizes Group-wise Correlation to compute matching costs between each view and other views.

Although many excellent anomaly detection methods and multi-view fusion methods are available now, an effective multi-view feature fusion method tailored for AD tasks needs to be designed explicitly. Therefore, we propose a novel approach, MVAD, for multi-view anomaly detection tasks, which significantly improves both effectiveness and efficiency.

\section{Method}
We propose a multi-view anomaly detection (MVAD) approach comprising a teacher-student model with a reverse knowledge distillation structure~\cite{rd4ad} and an intermediate multi-view feature fusion module, the Multi-View Adaptive Selection (MVAS) attention algorithm in Sec.~\ref{mvas}, at three different scales as shown in Fig. \ref{mvad}. MVAS has minimal $O((hw)^\frac{4}{3})$ computational complexity. Theoretical derivation and upper-bound validation  in Sec.~\ref{flopeq} demonstrate the efficiency of the proposed method.
\subsection{Preliminaries}
\label{defination}
\noindent
\textbf{Task Definition of multi-view AD.} Compared to the traditional anomaly detection input of $n \in \mathbb{N}$ batch-size images, the input for multi-view tasks is based on the number of \textit{samples}. Each input consists of $p \in \mathbb{N}$ samples, where each sample contains $v \in \mathbb{N}$ images from different views. Therefore, the actual input batch size is $p \times v \in \mathbb{N}$. During training, the features of each view $X_s \in \mathbb{R}^{p\times c\times h\times w}$ need to be fused with the features of the other views $X_m \in \mathbb{R}^{p\times (v-1)c\times h\times w}$ to obtain the enhanced features $Y_{s}^o \in \mathbb{R}^{p\times c\times h\times w}$ of the current view. During testing, the anomaly map $S_{px} \in \mathbb{R}^{pv\times H\times W}$ obtained serves as pixel-level anomaly scores. The maximum value of the entire anomaly map is taken as the image-level anomaly score $S_{im} \in \mathbb{R}^{pv}$. Finally, the maximum image-level anomaly score $S_{im}$ of the five views in each sample is taken as the $S_{sa} \in \mathbb{R}^{p}$ sample-level anomaly score. For sample-level GroundTruth $G_{sa} \in \mathbb{R}^{p}$, if any view within a sample contains an anomaly, then the sample is considered anomalous. Conversely, the sample is considered normal if no abnormal regions occur in any view of it.

\noindent
\textbf{Attention.} For input queries $Q \in \mathbb{R}^{N_q \times C}$, key $K \in \mathbb{R}^{N_k \times C}$, and value $V \in \mathbb{R}^{N_v \times C}$, the weights between the queries and keys are calculated by scaled dot-product attention. The weighted sum of the values is calculated by: 
\begin{equation}
\operatorname{Attention}(\mathbf{Q}, \mathbf{K}, \mathbf{V})=\operatorname{softmax}\left(\frac{\mathbf{Q} \mathbf{K}^T}{\sqrt{d_k}}\right) \mathbf{V},
\end{equation}
where $\sqrt{d_k}$ is used to avoid excessively large values in the result of the softmax function and to prevent gradient vanishing. Cross-attention involves queries from one input sequence and keys and values from another, making it conducive to multi-view feature integration.


\subsection{Multi-View Adaptive Selection}
\label{mvas}
To facilitate multi-view anomaly detection and fusion while minimizing computational complexity and memory usage, we propose a Multi-View Adaptive Selection (MVAS) attention algorithm. Detailed descriptions will be provided in the following sections, and the entire algorithm is presented in the form of PyTorch-like pseudo-code in Algorithm \ref{algorithm}. 

\noindent
\textbf{Neighborhood Attention Window.} Given an input multi-view feature map $\mathbf{X}_i \in \mathbb{R}^{v\times h\times w\times c}$, where $v$ denotes the number of views, the feature map is initially partitioned into neighbourhood attention windows of size $a \times a$ referred to as $\mathbf{X}_a \in \mathbb{R}^{v\times a^2 \times \frac{hw}{a^2} \times c}$, each window encompassing $\frac{hw}{a^2}$ feature vectors. A linear mapping is applied to feature map $\mathbf{X}_s \in \mathbb{R}^{ a^2 \times \frac{hw}{a^2} \times c}$ from a specific view to generate Query, while linear mappings are conducted on features $\mathbf{X}_m \in \mathbb{R}^{(v-1) \times a^2 \times \frac{hw}{a^2} \times c}$ from other views to produce Key and Value.
\begin{equation}
\mathbf{Q_s}=\mathbf{X}_s \mathbf{W}^q, \quad \mathbf{K}_m=\mathbf{X}_m \mathbf{W}^k, \quad \mathbf{V}_m=\mathbf{X}_m \mathbf{W}^v,
\end{equation}
where $\mathbf{W}^q, \mathbf{W}^k, \mathbf{W}^v \in \mathbb{R}^{c \times c}$ are linear mapping weights.

\noindent
\textbf{Multi-View Windows Adaptive Selection.} Subsequently, the objective is to identify the most correlated windows between the current view's feature window and those from other views to obtain a correlation matrix. Specifically, based on $\mathbf{X}_s \in \mathbb{R}^{ a^2 \times \frac{hw}{a^2} \times c}$, the partitioned window features for a single view $\mathbf{A}_s \in \mathbb{R}^{ a^2 \times c}$ are further derived, along with multi-view window features $\mathbf{A}_m \in \mathbb{R}^{ (v-1)a^2 \times c}$ from $\mathbf{X}_m \in \mathbb{R}^{(v-1) \times a^2 \times \frac{hw}{a^2} \times c}$. Then, the correlation matrix $\mathbf{A}_c \in \mathbb{R}^{a^2 \times (v-1)a^2}$ can be computed using the following formula:
\begin{equation}
\mathbf{A}_c = \mathbf{A}_s (\mathbf{A}_m)^K,
\end{equation}
The correlation matrix $\mathbf{A}_c$ reveals the semantic correlations between single-view window features and multi-view window features. Once the correlation matrix is obtained, the aim is to calculate the top-$k$ windows in which each window feature of the single view has the closest semantic correlation with the features of the multi-view windows. The ultimate objective is to derive the index matrix $\mathbf{I}_{m}^K \in \mathbb{R}^{a^2 \times k}$ of the highest-correlated multi-view windows:
\begin{equation}
\mathbf{I}_{m}^K = \operatorname{TopK\_Index} (\mathbf{A}_c),
\end{equation}
The features of each window in the current view are adaptively computed for semantic similarity with all windows from other views, selecting the top-$k$ most correlative windows to be the focus of subsequent attention computation.

\noindent
\textbf{Neighbourhood Correlative Cross-Attention.} To compute the cross attention of the feature map of a single view toward the top-$k$ most correlated neighbourhood windows of feature maps from other views, it is necessary to obtain the top-$k$ most correlated neighbourhood windows Key and Value $\mathbf{K}_{m}^K, \mathbf{V}_{m}^K \in \mathbb{R}^{a^2 \times \frac{khw}{a^2} \times c}$ by applying the most correlative index matrix $\mathbf{I}_{m}^K$ to the multi-view feature tensors $\mathbf{K}_m$ and $\mathbf{V}_m$. The cross-attention is applied to the input Query to get the enhanced multi-view feature fusion output $\mathbf{X}_{s}^o$:
\begin{equation}
\mathbf{X}_{s}^o=\operatorname{Attention}(\mathbf{Q}_s, \mathbf{K}_{m}^K, \mathbf{V}_{m}^K),
\end{equation}
Finally, the enhanced single view feature $\mathbf{X}_{s}^o$ should be transformed to the original input shape to get the unpatched output single view feature $\mathbf{Y}_{s}^o \in \mathbb{R}^{h\times w\times c}$. Iterate $v$ times, where $v$ represents the number of views in multi-view setting, and concatenate the results to obtain the final output $\mathbf{Y}_o \in \mathbb{R}^{v \times h\times w\times c}$.

\begin{algorithm}[t!]
	\caption{Pseudo-code of MVAS}
	\label{algorithm}
	\begin{algorithmic}[0]
		\Require\
		\State $X_i \in \mathbb{R}^{v\times h\times w\times c}$ is the multi-view input features
		\State $a \in \mathbb{N}$ is the size of the neighbourhood attention window 
		\State $k \in \mathbb{N}$ is the number of multi-view windows that need to be attended
		\Ensure\
		\State $Y_o \in \mathbb{R}^{v\times h\times w\times c}$ is the enhanced multi-view output features
	\end{algorithmic}
	\begin{algorithmic}[1]
            \State $Y_o=[]$
            \For {$j$ in range($X_i$.size(0))}
            \State $X_s = \operatorname{patch}(X_i[j], \operatorname{patchsize}=h // a)$
            \State $X_m = \operatorname{cat}(X_i[:j], X_i[j+1:],  \operatorname{dim=0})$
            \State $X_m = \operatorname{patch}(X_m), \operatorname{patchsize}=h // a) $
            \State $bs = X_m.\operatorname{shape}[0]\times X_m.\operatorname{shape}[1] $
            \State $X_m = X_m.\operatorname{view}(bs, X_m.\operatorname{shape}[2:])$
            \State $Q_s = \operatorname{linear_q}(X_s)$
            \State $K_m, V_m = \operatorname{linear_{kv}}(X_m).\operatorname{chunk(2, dim=-1)}$
		\State $A_s, A_m = Q_s.\operatorname{mean(dim=1)}, K_m.\operatorname{mean(dim=1)}$
            \State $A_c = A_s \cdot (A_m.\operatorname{transpose(-1,-2)})$
		\State $I_{m}^K = \operatorname{TopK\_Index}(A_c, k)$
            \State $K_{m}^K, V_{m}^K = \operatorname{gather}(K_m, I_{m}^K), \operatorname{gather}(V_m, I_{m}^K)$
            \State ${Y}_{s}^o = \operatorname{Attention}(Q_s, K_{m}^K, V_{m}^K)$
		\State  $Y_o.\operatorname{append}(\operatorname{unpatch}({Y}_{s}^o,a).\operatorname{unsqueeze(dim=0)})$
            \EndFor
            \State $Y_o = \operatorname{cat}(Y_o,\operatorname{dim=.0})$
            \State
            \Return $Y_o$
	\end{algorithmic}
\end{algorithm}
\subsection{Computation Complexity Analysis of MVAS}
\label{flopeq}
\noindent\textbf{Cross-Attention's Complexity.} 
The computational complexity $\Omega$ of cross-attention, where the single-view input is used as the query $Q \in \mathbb{R}^{(hw)\times c}$ and the multi-view inputs are used as the keys and values $K,V \in \mathbb{R}^{((v-1)hw)\times c}$, consists of two main components: \textit{Linear projection} and the \textit{Attention mechanism}.
\begin{equation}
\label{cross}
\begin{aligned}
& \Omega(\operatorname{Cross-Attention}) = \Omega(\operatorname{Proj}) + \Omega(\operatorname{C-Atten}) \\
& ~~~~~~~~~~~~~~~~~~~~~~~~= 3hwc^2 + 2(v-1)(hw)^2c \\
& ~~~~~~~~~~~~~~~~~~~~~~~~\approx O((hw)^2),
\end{aligned}
\end{equation}
where $v$ denotes the total number of views, $v - 1$ represents the number of views excluding the current one, $c$ is the feature dimension, and $hw$ is the spatial size of the feature map.
Cross-attention exhibits a high computational complexity of $O((hw)^2)$, which leads to significant consumption of GPU memory during training and inference.

\begin{table*}[t]

  \centering
  \caption{Comparison results on the Real-IAD~\cite{realiad} multi-view anomaly detection dataset under \textbf{multi-class settings} with SoTA methods, across \textbf{10 metrics} at \textbf{sample/image/pixel-level}. For easy comparison, the \textbf{Average} result is the mean of the former 10 results.}
\label{tab:compsota}%
  \renewcommand{\arraystretch}{1}
    \setlength\tabcolsep{3.0pt}
  \resizebox{1\linewidth}{!}{
    \begin{tabular}{cccccccccccc}
    \toprule
    \multicolumn{1}{c}{\multirow{2}[4]{*}{Method}} & \multicolumn{3}{c}{Sample-level} & \multicolumn{3}{c}{Image-level} & \multicolumn{4}{c}{Pixel-level} & \multirow{2}[4]{*}{Average} \\
\cmidrule(r){2-4} \cmidrule{5-7} \cmidrule(l){8-11}  \multicolumn{1}{c}{} & AU-ROC & AP    & F1-max & AU-ROC & AP    & F1-max & AU-ROC & AP    & F1-max & AU-PRO &  \\
    \midrule
    Simplenet~\cite{simplenet} & 61.2  & 77.9  & 82.1  & 82.6  & 79.2  & 74.1  & 76.3  & 3.4   & 6.9   & 42.5  & 58.6  \\
    CFA~\cite{cfa}   & 65.8  & 80.2  & 82.2  & 58.3  & 53.9  & 61.7  & 82.4  & 2.2   & 5.9   & 51.2  & 54.4  \\
    CFLOW-AD~\cite{cflow} & 80.2  & 89.8  & 84.9  & 75.1  & 74.1  & 69.1  & 95.0  & 18.1  & 21.7  & 81.7  & 69.0  \\
    PyramidalFlow~\cite{pyramidflow} &   63.1    &   79.3    &   82.0    &   56.2    &   52.6    &    62.5   &     71.5  &   1.5    &   1.7    &    35.5   & 50.6 \\
    RD4AD~\cite{rd4ad}    & 85.7  & 92.4  & 87.7  & 83.0  & 79.6  & 74.4  & 97.3  & 25.8  & 33.5  & \underline{90.4}  & 75.0 \\
    ViTAD~\cite{vitad} & 88.8  & 94.3  & \underline{89.3}  & 82.8  & 80.0  & 73.7  & 97.1  & 23.9  & 31.8  & 84.3  & 74.6  \\
    UniAD~\cite{uniad} & 87.3  & 93.1  & 88.9  & 83.0  & 83.4  & 74.6  & 97.4  & 23.7  & 31.4  & 87.3  & 74.9  \\
    RD++~\cite{rd++}  & 87.0  & 93.3  & 88.2  & \underline{83.7}  & \underline{80.9}  & \underline{74.9}  & \underline{97.7}  & 25.8  & 33.5  & 90.0  & \underline{75.5}  \\
    DesTSeg~\cite{destseg} & \underline{89.0}  & \underline{94.6}  & 89.2  & 82.3  & 79.2  & 73.2  & 94.6  & \textbf{37.9}  & \textbf{41.7}  & 40.6  & 72.2  \\
    Ours  & \textbf{90.2}  & \textbf{95.3}  & \textbf{90.1}  & \textbf{86.6}  & \textbf{84.8}  & \textbf{77.2}  & \textbf{97.9}  & \underline{30.3}  & \underline{36.8}  & \textbf{91.2}  & \textbf{78.0}  \\
    \bottomrule
    \end{tabular}%
    }
\end{table*}
\noindent \textbf{Ours' Complexity.} Therefore, MVAS is proposed with less computational complexity, which consists of three parts: the Neighbourhood Attention Window, the Multi-View Windows Adaptive Selection and Neighbourhood Correlative Cross-Attention. 
The FLOPs calculation for the Neighbourhood Attention Window primarily includes the linear projection of $QKV$. The FLOPs for Multi-View Windows Adaptive Selection include the computation of the correlation matrix and the Top-$k$ selection. Finally, The Neighbourhood Correlative Cross-Attention is a pure cross-attention mechanism.
The total computational complexity is as follows:
\begin{equation}
\label{ours}
\begin{aligned}
& \Omega(\operatorname{Ours}) = \Omega(\operatorname{NAW}) + \Omega(\operatorname{MVWAS}) + \Omega(\operatorname{NCCA}) \\
&~~~~~~~~~~=\Omega(\operatorname{Proj}) + \Omega(\operatorname{Cor}) +  \Omega(\operatorname{Topk}) + \Omega(\operatorname{C-Atten}) \\
& ~~~~~~~~~~= 3hwc^2 + 2(v-1)(a^2)^2c + 2k\frac{(hw)^2}{a^2}c\\
& ~~~~~~~~~~= 3hwc^2 + c(2(v-1)a^4 + k\frac{(hw)^2}{a^2} + k\frac{(hw)^2}{a^2}) \\
& ~~~~~~~~~~\geq 3hwc^2 +  3c(2(v-1)a^4 \cdot k\frac{(hw)^2}{a^2} \cdot k\frac{(hw)^2}{a^2})^\frac{1}{3} \\
& ~~~~~~~~~~= 3hwc^2 + 3c(2(v-1)k)^\frac{1}{3}(hw)^\frac{4}{3} \\
& ~~~~~~~~~~\approx O((hw)^\frac{4}{3}),
\end{aligned}
\end{equation}
\noindent where $1\leq a\leq \sqrt{hw}$ is the neighbourhood attention window size and $1\leq k\leq (v-1)\frac{hw}{a^2}$ is the number of the most correlated neighbourhood windows. The inequality between arithmetic and geometric means is applied here. The equality in Eq. \ref{ours} holds if and only if: 
\begin{equation}
\label{min}
2(v-1)a^4 = k\frac{(hw)^2}{a^2} \quad \Rightarrow \quad \frac{a^6}{k} = \frac{(hw)^2}{2(v-1)} ,
\end{equation}
Therefore, MVAS could achieve approximately $O((hw)^\frac{4}{3})$ computational complexity by adjusting the neighbourhood attention window size.

\noindent \textbf{Upper Bound Analysis of Ours' Complexity.} In the extreme case where $a = \sqrt{hw}$, the entire feature map is treated as a single window. Simultaneously, let $k = (v-1)\frac{hw}{a^2}$, selecting all other viewpoint images as the Top-$k$ for the correlation matrix and the $KV$ for cross-attention computation. The maximum computational complexity in this scenario is:
\begin{equation}
\label{max}
\begin{aligned}
& \Omega(\operatorname{Ours})_{max} = 3hwc^2 + 2(v-1)(a^2)^2c + 2k\frac{(hw)^2}{a^2}c\\
& ~~~~~~= 3hwc^2 + 2(v-1)(hw)^2c + 2(v-1)\frac{hw}{a^2}\frac{(hw)^2}{a^2}c\\
& ~~~~~~= 3hwc^2 + 2(v-1)(hw)^2c + 2(v-1)(hw)c,\\
& ~~~~~~= \Omega(\operatorname{Cross-Attention}) + 2(v-1)(hw)c,\\
\end{aligned}
\end{equation}
Compared to the direct computation of Cross-Attention, the computational complexity in the extreme case increases by an additional term $2(v-1)(hw)c$. This term originates from the Multi-View Windows Adaptive Selection process. However, since all windows from all views are selected under this extreme condition, this operation is essentially redundant and only introduces extra computational overhead. Nonetheless, the impact of this additional term on the overall computational complexity is relatively minor compared to the preceding term $2(v-1)(hw)^2c$.

\begin{figure*}[b!]
\centering
\includegraphics[width=1\textwidth]{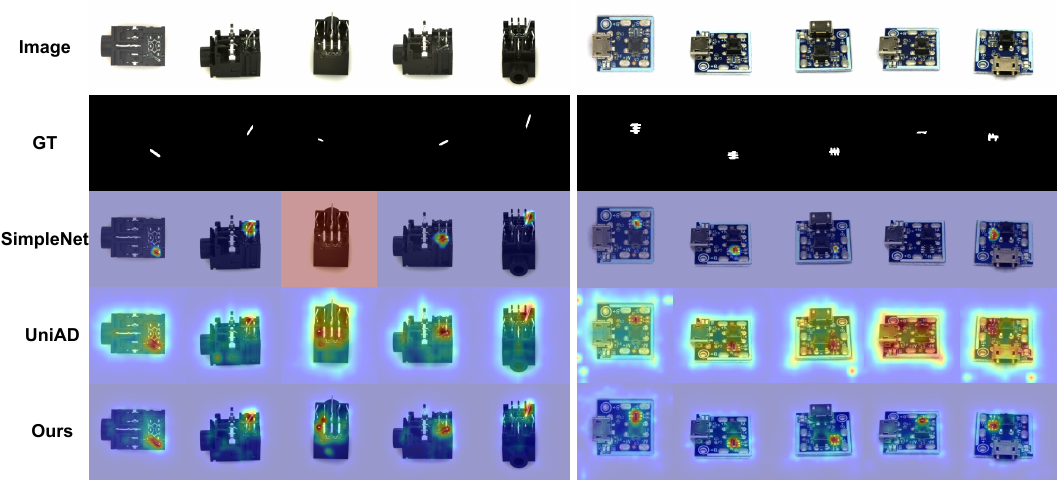}
\caption{Qualitative visualized results for pixel-level anomaly segmentation on two complex examples audiojack ({\textit{Left}}) and PCB (\textit{Right}).}
\label{visual}
\end{figure*}

\begin{table*}[htbp]
  \centering
  \caption{Comparison of \textbf{hyperparameters} and \textbf{efficiency} with mainstream SoTA methods on Real-IAD~\cite{realiad} dataset.  \textbf{TP}: \textbf{Throughput} on a NVIDIA A100 GPU. Performance refers to the \textbf{average value of 10 metrics} across three levels: sample, image, and pixel.}
    \label{tab:eff}%
  \renewcommand{\arraystretch}{1}
    \setlength\tabcolsep{3.0pt}
  \resizebox{1\linewidth}{!}{
    \begin{tabular}{ccccccccccc}
    \toprule
    \multirow{2}[4]{*}{Method} & \multicolumn{4}{c}{Hyper Paramters} & \multicolumn{5}{c}{Efficiency} & \multirow{2}[4]{*}{Performance} \\
\cmidrule(r){2-5} \cmidrule(l){6-10}  & BS & Optim & LR    & Backbone & Params & FLOPs & Train Memory & Train Time & TP \\
    \midrule
    Simplenet~\cite{simplenet} & 40    & AdamW & 1e-4  & WideResNet-50 & 72.8M & 88.6G & 7,576M & 41h   & 116.4  & 58.6 \\
    CFA~\cite{cfa}   & 40    & AdamW & 1e-3  & WideResNet-50 & 38.6M & 276.5G & 19,538M & 32h   & 181.6 & 54.4 \\
    CFLOW-AD~\cite{cflow} & 40    & Adam  & 2e-4  & WideResNet-50 & 237M  & 143.5G & \underline{4,924M} & 120h  & 102.3 & 69.0 \\
    PyramidalFlow~\cite{pyramidflow} & 10    & Adam  & 2e-4  & ResNet-18 & 34.3M & 4810G & 14,314M & 131h  & 82.6  & 50.6 \\
    RD4AD~\cite{rd4ad}    &  40   & Adam  & 5e-3  & WideResNet-50 & 80.6M & 142.0G & 11,614M & \underline{12h}   & 231.2  & 75.0\\
    ViTAD~\cite{vitad} & 40    & AdamW & 1e-4  & ViT-S-p16 & 39.0M & 48.5G & \textbf{4,842M} & \underline{12h}   & \textbf{468.9}  & 74.6 \\
    UniAD~\cite{uniad} & 40    & AdamW & 1e-4  & EfficientNet-b4 & \underline{24.5M} & \textbf{16.8G} & 7,324M & \underline{12h}   & \underline{427.4} & 74.9 \\
    RD++~\cite{rd++}  & 40    & Adam  & 1e-3  & WideResNet-50 & 96.1M & 187.5G & 17,072M & 124h  & 159.2 & \underline{75.5} \\
    DesTSeg~\cite{destseg} & 40    & SGD   & 0.4 & ResNet-18 & 35.2M & 153.0G & 11,521M & 15h   & 221.2  &  72.2 \\
    Ours  & 40    & AdamW & 5e-3  & ResNet-34 & \textbf{18.4M} & \underline{45.0G} & 6,744M & \textbf{11h}  & 349.4 & \textbf{78.0} \\
    \bottomrule
    \end{tabular}}
\end{table*}%

\vspace{-0.5em}
\subsection{Training and Testing of MVAD}
For each $j\text{-}th$ scale, there are several corresponding $\operatorname{MVAS^j}$ blocks with the same number of dimensions. Each single-view feature, after incorporating positional encoding $\mathbb P$, is fed into the MVAS block, which is defined as:
\begin{equation}
\begin{aligned}
& X_{o}^j=\operatorname{LN^j}\left(\operatorname{MVAS^j}\left(X_{i}^j + \mathbb P \right)\right)+X_{i}^j, \\
& X_{o}^j=\operatorname{LN^j}\left(\operatorname{MLP^j}\left(X_{o}^j\right)\right)+X_{o}^j,
\end{aligned}
\end{equation}
where $X_{o}^j$ is the enhanced multi-view features for $j\text{-}th$ scale. Then, the enhanced features at each scale are integrated using an FPN-like structure and fed into the decoder model. Finally, the loss function is computed by summing the MSE loss at each scale of the encoder $f_{\mathcal{E}}^j$ and decoder ${f_{\mathcal{D}}^j}$:
\begin{equation}
\mathcal{L}=\sum_{j \in J}\{\frac{1}{H\times W}\|f_{\mathcal{E}}^j - {f_{\mathcal{D}}^j}\|_{2}^2\},
\end{equation}
where $J$ is the number of feature stages used in experiments.

During the testing phase, following~\cite{rd4ad} we utilize cosine similarity to calculate pixel-level anomaly scores for the encoders and decoders across three different stages. The maximum value of the anomaly map is taken directly as the image-level anomaly score. The maximum anomaly score among images of the same sample from different views is taken as the sample-level anomaly score.

\section{Experiments}
\subsection{Datasets and Implementations Details}
\noindent \textbf{Real-IAD Dataset.} 
The Real-IAD~\cite{realiad} dataset is a large-scale multi-view anomaly detection dataset collected from real-world industrial components. It includes 30 classes of components, each captured from five different views. The dataset comprises 99,721 normal and 51,329 anomalous samples, totaling 150K images for training and testing. Images are captured with a professional camera at ultra-high resolutions of $2K\sim5K$. Pixel-level segmentation masks and multi-view image labels are provided for evaluating multi-view anomaly detection.

\noindent \textbf{Task setting.} We conducted experiments on multi-class settings in the Real-IAD dataset. In the multi-class setting, all categories are trained and tested by a single model. 

\noindent \textbf{Evaluation Metric.} The Area Under the Receiver Operating Characteristic Curve (AU-ROC), Average Precision (AP), and F1-score-max (F1-max) are simultaneously used for the pixel-level, image-level, and sample-level evaluations. Area Under the Per-Region-Overlap (AU-PRO) is used for anomaly localization at the pixel level.

\noindent\textbf{Training and Testing Setups.} All methods use $n$ samples, each with $5$ views, resulting in a batch size of $5n$. Following the unsupervised anomaly detection paradigm, all methods are trained on normal samples and tested on both normal and anomaly samples. All methods obtain anomaly localization scores for each image, with the maximum score used as the image-level anomaly detection score. Both our method and the comparison methods use the maximum anomaly detection score among all view images of the same sample as the sample-level anomaly score. Thus, all methods and ours share the same training and testing strategies. The only difference is that comparison methods treat different views as separate batches without inter-view information interaction, while our method, using the MVAS algorithm, integrates features from different views within the batch.

\noindent \textbf{Implementation Details.}
All input images are resized to $256\times256$ without additional data augmentation. For multi-view tasks, training is conducted using all available view images $v=5$, while the input samples are $8$, resulting in a batch size of $40$. A pre-trained ResNet34 model serves as the teacher feature extractor, with a corresponding reverse distillation Re-ResNet34 model employed as the student model for training. AdamW optimizer is utilized with a decay rate of 1e-4 and an initial learning rate of $0.005$. The model is trained for $100$ epochs for multi-class settings on a single NVIDIA A100 40GB. For a balance between effectiveness and efficiency, we set $\{a_1,a_2,a_3\}=$\{14,9,6\}, $\{k_1,k_2,k_3\}=\{4,4,4\}$, and $\{N_1,N_2,N_3\}=\{1,2,4\}$ for the size of the neighbourhood attention window, the top-$k$ selection, and the number of MVAS blocks for each stage. 
The selection of $a$ and $k$ is further discussed in the ablation study (\textit{cf.} Sec.\ref{ablation}).

\begin{table}[t]
  \centering
  \caption{Ablation studies on the \textbf{window size $a$} and \textbf{top $k$} in MVAS at different stages. Sample/Image/Pixel denotes the \textbf{average values} of metrics at their respective levels. This kind of display continues for all subsequent experiments.}
\label{abl:kw}%
  \renewcommand{\arraystretch}{1.2}
    \setlength\tabcolsep{2.0pt}
  \resizebox{1\linewidth}{!}{
    \begin{tabular}{cccccc}
    \toprule
    Index & $\{k_1,k_2,k_3\}$     & $\{a_1,a_2,a_3\}$     & FLOPs & TP & Sample/Image/Pixel \\
    \midrule
    \ding{172}     & $\{1,1,1\}$ & $\{11,7,4\}$ & \textbf{44.5G}  & \textbf{368.7}  & \textbf{92.0}/\textbf{82.9}/\underline{63.7} \\
    \ding{173}     & $\{2,2,2\}$ & $\{13,8,5\}$ & \underline{44.7G}  & \underline{353.4}  & 91.8/\underline{82.8}/\underline{63.7} \\
    \ding{174}     & $\{4,4,4\}$ & $\{14,9,6\}$ & 45.0G    & 349.4  & \underline{91.9}/\textbf{82.9}/\textbf{64.1} \\
    \ding{175}     & $\{8,8,8\}$ & $\{16,10,6\}$ &45.5G & 339.1  & 91.7/82.4/63.2 \\
    \ding{176}     & $\{8,16,32\}$ & $\{16,11,8\}$ & 45.6G  & 331.6  & 91.7/82.7/\underline{63.7} \\
    \ding{177}     & $\{16,32,64\}$ & $\{18,13,9\}$ & 47.5G  & 228.5  & 91.8/\underline{82.8}/\underline{63.7} \\
    \ding{178}     & $\{16,32,64\}$ & $\{8,8,8\}$ & 49.1G  & 223.4  & 91.7/82.4/63.4 \\
    \bottomrule
    \end{tabular}%
    }
\end{table}%
\noindent
\subsection{Comparison with SoTAs on multi-view Real-IAD}
\label{result}
Because no method is specifically designed for multi-view AD, we selected \textit{1).} Augmentation-based SimpleNet~\cite{simplenet}, \textit{2).} Embedding-based CFA~\cite{cfa}, CFLOW-AD~\cite{cflow}, and PyramidFlow~\cite{pyramidflow}, \textit{3).} Reconstruction-based RD4AD~\cite{rd4ad}, and ViTAD~\cite{vitad} and \textit{4).} Hybrid Augmentation and Reconstruction-based UniAD~\cite{uniad}, RD++~\cite{rd++}, and DeSTSeg~\cite{destseg} as our comparative methods. Tab.~\ref{tab:compsota} presents the average results of all methods across 10 metrics for all categories under the multi-class settings.  Tab.~\ref{tab:eff} directly compares the hyperparameters, efficiency, and performance on multi-view anomaly detection Real-IAD datasets for various methods, employing a single NVIDIA A100 GPU.

\noindent \textbf{Quantitative Results.} Tab.~\ref{tab:compsota} presents the average results of our method compared to nine SoTA methods on the Real-IAD~\cite{realiad} multi-view anomaly detection dataset under multi-class settings. Each metric represents the average value across all categories. Augmentation-based SimpleNet~\cite{simplenet} and embedding-based CFA~\cite{cfa}, CFLOW-AD~\cite{cflow}, and PyramidFlow~\cite{pyramidflow} methods perform poorly on the Real-IAD dataset in multi-class settings, indicating limitations in these approaches. Reconstruction-based and hybrid methods achieve relatively better results, with RD++ obtaining the highest average score of 75.5 across all metrics. However, our method surpasses RD++~\cite{rd++} by an average of \textbf{+2.5$\uparrow$} across all metrics, reaching \textbf{78.0}, significantly outperforming other methods. Although the DeSTSeg~\cite{destseg} method achieves high segmentation accuracy due to its use of data augmentation for training the segmentation network, it suffers from lower AUPRO scores due to missed detections, resulting in a lower overall average metric. In contrast, our method achieves SoTA performance without requiring additional data augmentation for training.

\begin{table}[t]
  \centering
  \caption{Ablation studies on different backbones.}
  \label{abl:bb}
  \renewcommand{\arraystretch}{1.2}
  \resizebox{1\linewidth}{!}{
\begin{tabular}{ccccc}
    \toprule
    \multirow{2}[4]{*}{Backbone} & \multicolumn{3}{c}{Efficiency} & Effectiveness \\
\cmidrule(r){2-4}      & FLOPs & Params & TP & \multicolumn{1}{c}{Sample/Image/Pixel} \\
    \midrule
    ResNet18 & \textbf{24G}   & \textbf{10.9M} & \textbf{432.6}  & 83.2/73.7/57.8 \\
    ResNet34 & \underline{45G}  & \underline{18.4M} & \underline{349.4}  & 91.9/82.9/\textbf{64.1} \\
    ResNet50 & 146G  & 95.5M & 162.3  & \underline{92.3}/\textbf{83.1}/62.0 \\
    WideResNet50 & 216G  & 121M  & 134.9  & \textbf{92.7}/\underline{83.0}/\underline{62.8} \\
    \bottomrule
\end{tabular}}
\end{table}%
\noindent \textbf{Qualitative Results.}
To further visually demonstrate the effectiveness of our approach, we conducted qualitative experiments to illustrate the results of anomaly localization compared with SimpleNet and UniAD. The comparison focused on two classes of complex industrial components from five views of the same sample, as illustrated in Fig. \ref{visual}. SimpleNet fails to recognize tiny anomalies, which are depicted in red on the anomaly maps, in the audiojack category. UniAD contains more false positives in the PCB category. Our approach demonstrates high accuracy and low false alarm rate, showcasing strong multi-view anomaly localization capabilities. Comparisons across all other category samples and methods will be presented in the appendix. 

\begin{figure*}[tb]
\centering
\includegraphics[width=1\textwidth]{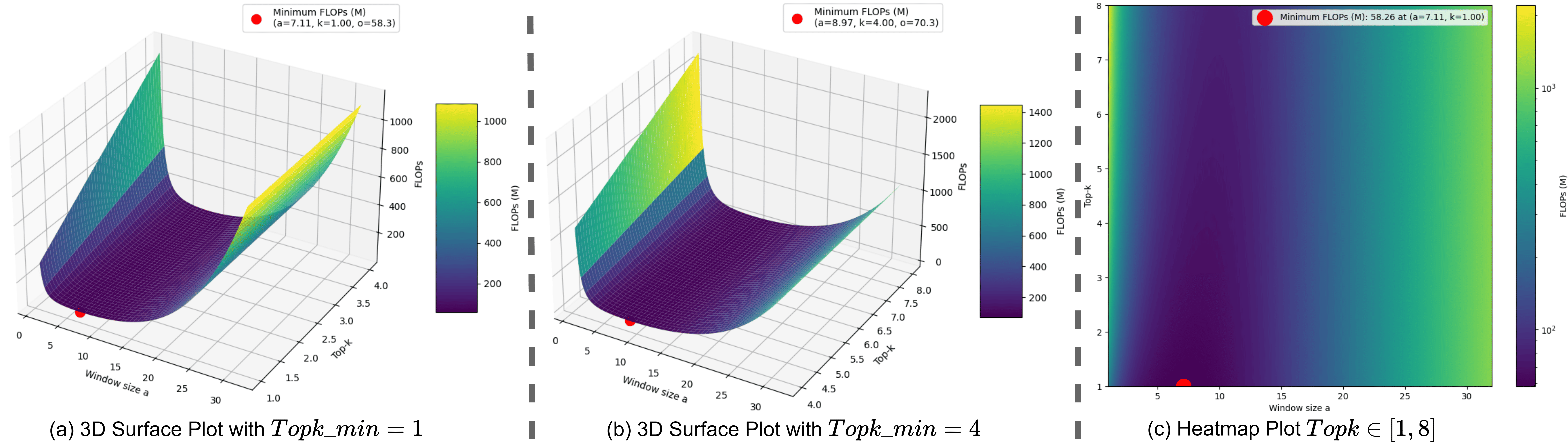}
\vspace{-1.5em}
\caption{The qualitative results of FLOPs with varying window size \(a\) and top-$k$. The feature map from the second stage is used as an example for calculations, with \(h=w=2^5\), \(c=2^7\), and \(v=5\). \textbf{(a)} 3D qualitative results when top-$k_{min}=1$. \textbf{(b)} 3D qualitative results when top-$k_{min}=4$. \textbf{(c)} Heatmap results when top-$k_{min}=1$. The red points indicate the points where FLOPs reach their minimum value, and the corresponding values of \(a\) and \(k\) are displayed.}
\label{fig:3d}
\end{figure*}

\subsection{Ablation Study}
\label{ablation}
\noindent
\textbf{Efficiency Comparison.} Tab.~\ref{tab:eff} presents the hyperparameters, efficiency, and performance of ours compared to all SoTA methods. For hyperparameters, we list batch size, optimizer, learning rate, and the feature extraction network used. To ensure a fair comparison of training memory usage and training time, we set the number of training epochs to 40 for all methods. However, for PyramidFlow~\cite{pyramidflow}, we set the epochs to 10 due to memory overflow issues at 40 epochs. The learning rate is linearly adjusted based on the original papers of each method.
For efficiency, we evaluate the number of parameters, FLOPs, training GPU memory usage, training time, and GPU throughput. Our method has the \textbf{smallest parameters} at only \textbf{18.4M}, the \textbf{shortest training time}, the \textbf{second-fewest FLOPs}, and the \textbf{third-lowest training memory usage}. The GPU throughput tested on a single Nvidia V100 is slightly lower than ViTAD~\cite{vitad} and UniAD~\cite{uniad} due to frequent reshape operations in the Multi-View Adaptive Selection process, which involves data copying and increases memory access overhead, thereby prolonging memory retrieval times. These factors result in reduced throughput for the proposed MVAS. We plan to further optimize the code to reduce the frequency of reshaping and improve model throughput.
Despite this, our method achieves an average score of \textbf{78.0} across all 10 evaluation metrics, surpassing the second-best method by \textbf{+2.5$\uparrow$} points. This demonstrates that our method achieves a superior trade-off between efficiency and performance, outperforming existing SoTA methods.

\begin{table}[t]
  \centering
  
  \caption{Ablation studies on different architecture.}
  \label{abl:archite}
  \renewcommand{\arraystretch}{1.0}
  \resizebox{1\linewidth}{!}{
\begin{tabular}{ccccc}
    \toprule
    \multirow{2}[4]{*}{Backbone} & \multicolumn{3}{c}{Efficiency} & Effectiveness \\
\cmidrule(r){2-4}      & Params & FLOPs & Memory & \multicolumn{1}{c}{Sample/Image/Pixel} \\
    \midrule
    UniAD   & 24.5M & 16.8G & 7324M                & 89.8/80.3/59.9          \\
+MVAS   & 25.6M & 18.7G & 7638M                &   \textbf{90.3/81.0/60.4}  \\
\midrule
ViTAD   & 39.0M & 48.5G & 4842M                & 90.8/78.8/59.3 \\
+MVAS   & 44.9M & 54.6G & 5092M                & \textbf{91.4/79.6/59.7}          \\
\midrule
RD-Base & 15.4M & 39.5G & 3120M                & 89.3/79.0/59.7          \\
+MVAS   & 18.4M & 45.0G & 6744M                & \textbf{91.9/82.9/64.1}          \\
    \bottomrule
\end{tabular}}
\end{table}%
\noindent\textbf{Quantitative Analysis of Window Size and Top-K Selection.}
From Eq.~\ref{ours}, the FLOPs of the proposed MVAS is $\Omega(\operatorname{MVAS}) = 3hwc^2 + 2(v-1)(a^2)^2c + 2k\frac{(hw)^2}{a^2}c$. Given a fixed backbone, input resolution, and the number of views, the only variables in this equation are window size \(a\) and top-$k$. It is straightforward to deduce from the equation that \textit{FLOPs are directly proportional to the number of top-$k$.} \textit{\textbf{1).}} the first step is to determine the number of top-$k$. As shown in the second column of Tab.~\ref{abl:kw}, top-$k$ ranges from 1 (minimum FLOPs) to 8, and then increases progressively across different stages. The progressive increase in top-$k$ across stages is due to the decreasing size of feature maps $(hw)$, which means that \textit{increasing top-$k$ in later stages introduces negligible FLOPs growth}. For example, the comparison between experiments \ding{176} and \ding{175} shows only a 0.1G increase in FLOPs.
\textit{\textbf{2).}} After determining the top-$k$, is to consider the window size \(a\). From Eq.~\ref{min}, we find that the minimum FLOPs occur when $\frac{a^6}{k} = \frac{(hw)^2}{2(v-1)}$. At this point, the window size \(a\) can be calculated based on the size of the feature maps at different stages, the number of views \(v\), and the current value of \(k\). The third column of Tab.~\ref{abl:kw} shows the rounded values of \(a\) calculated from the equation except for \ding{178}. These calculated values represent the minimum FLOPs for different \(k\), which is the optimal solution.
\textit{\textbf{3).}} From the results in Tab.~\ref{abl:kw} (\ding{172}-\ding{177}), we observe that as $k$ increases, the model's FLOPs increase while throughput decreases. However, \textit{the performance does not improve with higher top-$k$ values}. This is because \textit{increasing top-$k$ values introduce optimization difficulties and redundant information}. Experiment \ding{174}, with $k=4$, shows the best performance, which equals the number of other views in the multi-view setup. This indicates that \textit{selecting one window from each other view for a given view's window avoids redundancy and enhances the accuracy of the correlation matrix}. Comparing experiments \ding{177} and \ding{178}, \ding{177} uses the \(a\) calculated for minimum FLOPs from Eq.~\ref{min}, while \ding{178} uses a fixed value. \ding{177} achieves -1.6G$\downarrow$ lower FLOPs and improves the Sample/Image/Pixel metrics by +0.1$\uparrow$/+0.4$\uparrow$/+0.3$\uparrow$, respectively.
In conclusion, Tab.~\ref{abl:kw} demonstrates the effectiveness of our method in selecting the optimal window size \(a\) and top-$k$ using Eq.~\ref{ours} and Eq.~\ref{min}.

\begin{figure}[!t]
\centering 
\includegraphics[width=2.7in]{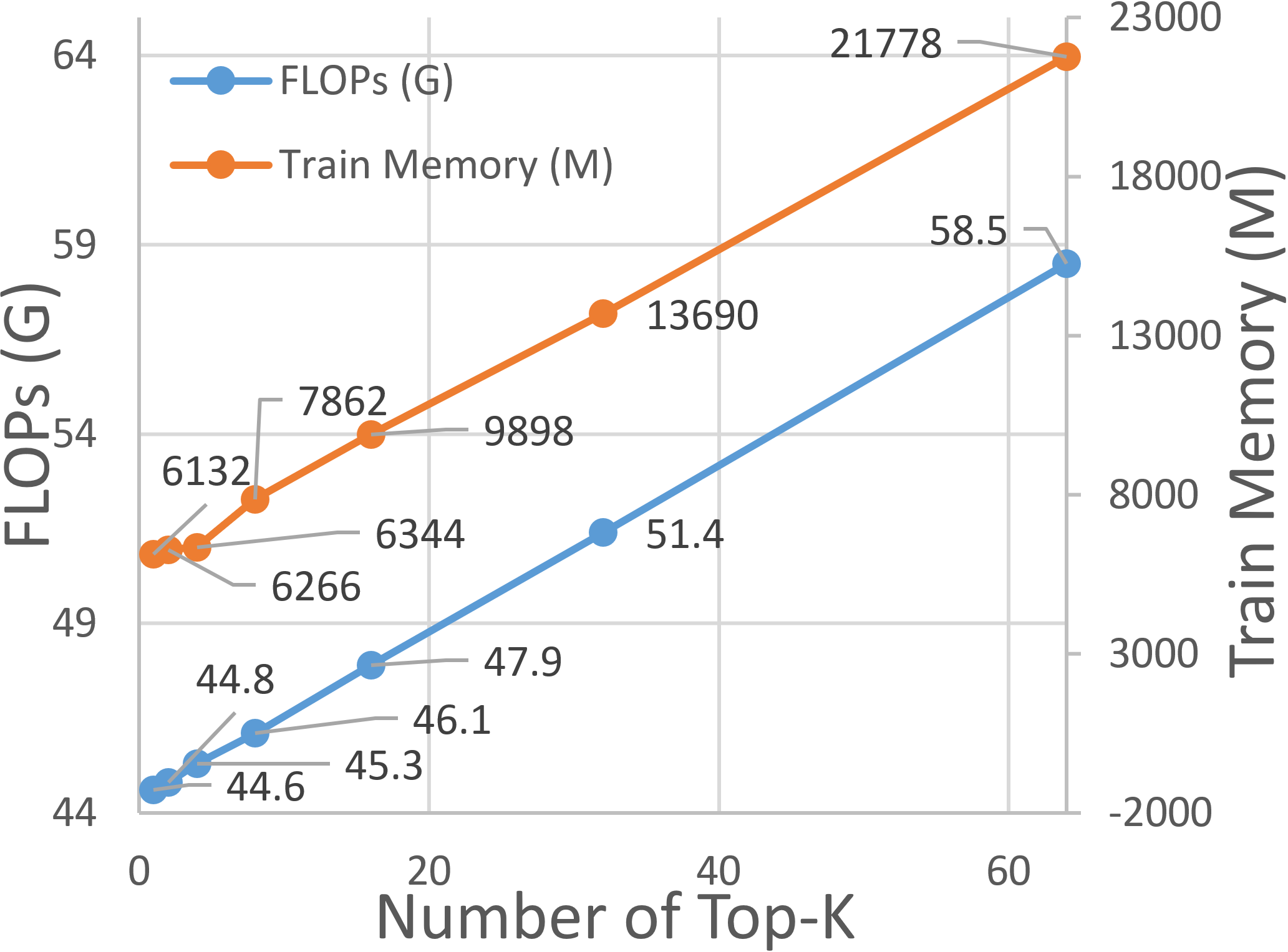}
\caption{Ablation studies on top-$k$ with $a=8$} 
\label{fig:topk}
\end{figure}
\noindent\textbf{Qualitative Analysis of Window Size and Top-K Selection.} Fig.~\ref{fig:3d} presents a qualitative experimental analysis of FLOPs with respect to changes in \(a\) and \(k\). To clearly illustrate these variations, we use (a) and (b) for 3D scatter plots with different minimum top-$k$ values, and (c) for a heatmap. FLOPs are calculated using Eq.~\ref{ours}, with parameters from the second stage: \(h=w=2^5\), \(c=2^7\), and \(v=5\). 
From the three figures, it is observed that the minimum FLOPs occur at the smallest top-$k$ values, and the corresponding \(a\) values align with those calculated using Eq.~\ref{ours} for the minimum top-$k$. Specifically, in Fig.~\ref{fig:3d} (a) and (c), the minimum is achieved at \(k=1\) and \(a=7.11\), consistent with \(a_2=7\) in \ding{172} of Tab.~\ref{abl:kw}. In Fig.~\ref{fig:3d} (b), the minimum is achieved at \(k=4\) and \(a=8.97\), consistent with \(a_2=9\) in \ding{174} of Tab.~\ref{abl:kw}. 
Additionally, the analysis indicates that \textit{for a fixed \(k\), both excessively large and small values of \(a\) lead to a significant increase in FLOPs.} Therefore, the qualitative experimental results further validate the theoretical consistency and the effectiveness of the method.

\noindent
\textbf{The Efficiency and Effectiveness of Different Pre-trained Backbone.} We investigate the impact of different pre-trained backbones, including ResNet18, ResNet34, ResNet50, and WideResNet50 in Tab.~\ref{abl:bb}. 
As the backbone network deepens, the number of channels extracted at various stages increases, leading to more model parameters and FLOPs, and reduced throughput. ResNet18, while efficient, performs the worst. WideResNet50 excels in sample-level metrics, ResNet50 in image-level metrics, and ResNet34 in pixel-level metrics. This suggests deeper networks better extract macro-level semantics but lose micro-level details. We adopt ResNet34 for the best trade-off between efficiency and effectiveness.

\begin{figure}[!t]
\centering 
\includegraphics[width=2.7in]{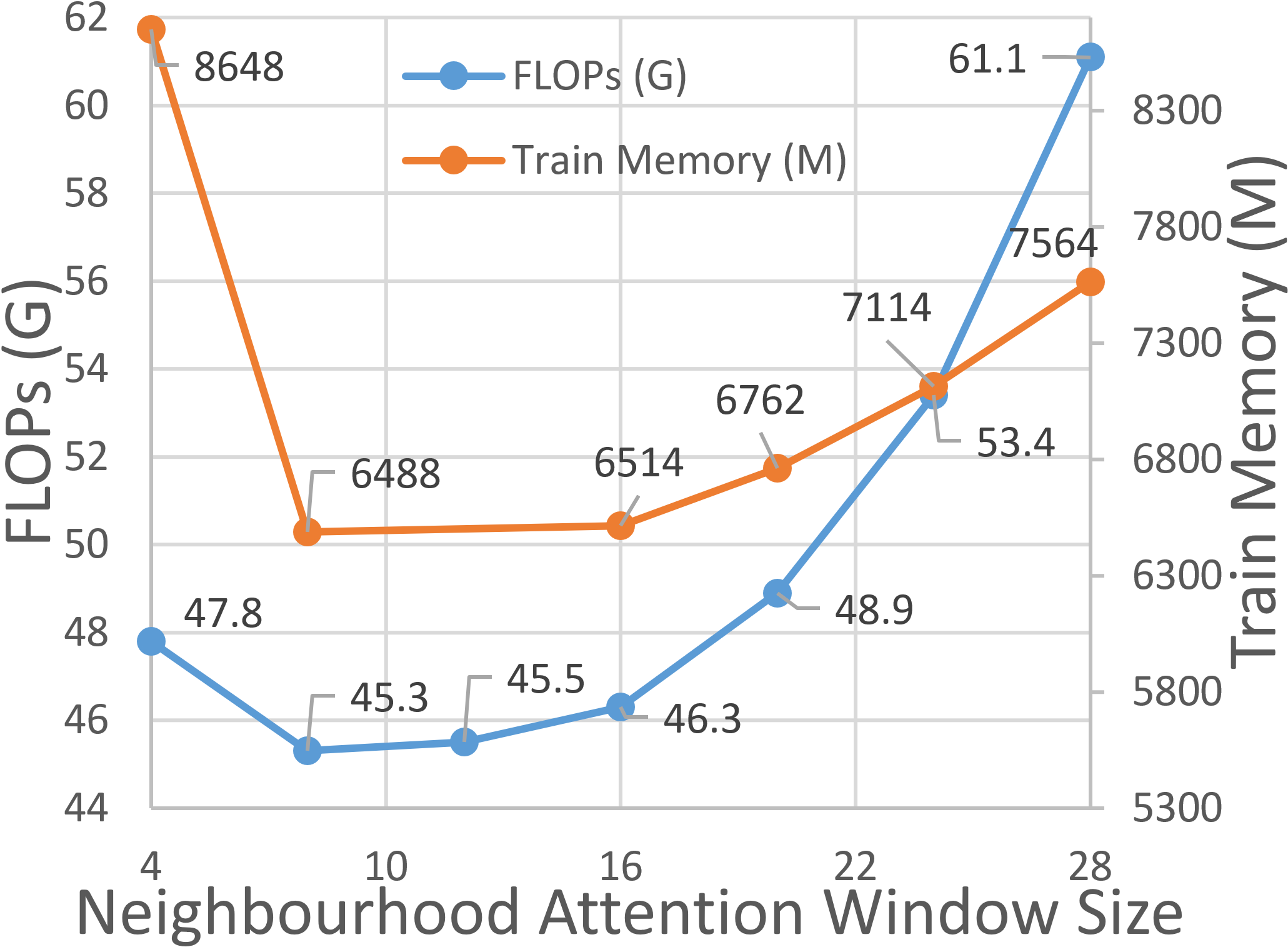}
\caption{Ablation studies on window size with $k=4$} 
\label{fig:ak}
\end{figure}
\noindent\textbf{Control Variable Experiment on Window Size and Top-K.} Fig.~\ref{fig:topk} and~\ref{fig:ak} respectively show the impact on FLOPs and train memory when varying top-$k$ with the fixed window size $a$, and when varying window size $a$ with the fixed top-$k$. Fig.~\ref{fig:topk} demonstrates that \textit{both FLOPs and train memory increase linearly with the increase of top-$k$, consistent with the theory in Eq.~\ref{ours}}. Fig.~\ref{fig:ak} shows that \textit{with the fixed top-$k$, increasing initially decreases and then increases FLOPs and train memory}. This is because \textit{FLOPs reach a minimum when a satisfies Eq.~\ref{min}, while in other cases, FLOPs increase, further validating the previous theoretical findings.}

\begin{figure}[t]
\centering
\includegraphics[width=0.5\textwidth]{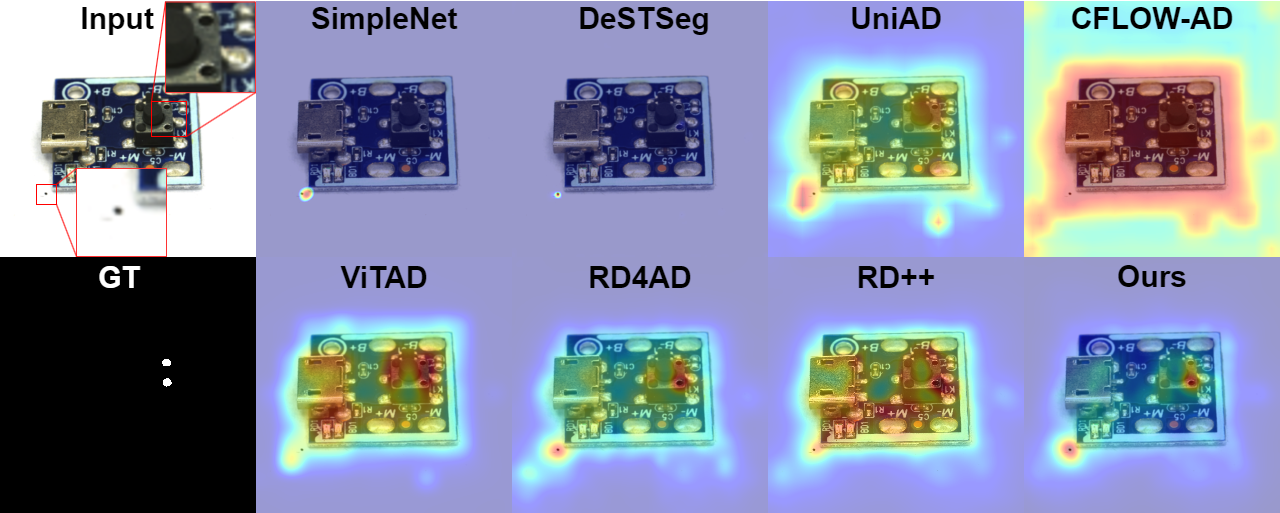}
\vspace{-1.5em}
\caption{\textbf{Failure Case Analysis.} The top left corner displays the input image, featuring two minor defects and a background impurity. The ground truth marks the positions of the two small defects on the right. The eight anomaly score maps on the right compare our method with seven SoTA methods.}
\label{fig:fail}
\end{figure}

\noindent\textbf{Failure Case Analysis.} Fig.~\ref{fig:fail} illustrates a challenging failure case with seven SoTA methods. All methods misidentify the background impurity as a defect. SimpleNet~\cite{simplenet} and DeSTSeg~\cite{destseg} fail to detect the actual small defects, focusing solely on the impurity. CFLOW-AD~\cite{cflow}, UniAD~\cite{uniad}, ViTAD~\cite{vitad}, and RD++~\cite{rd++} exhibit abnormal scores across most object regions, resulting in numerous false positives. RD4AD~\cite{rd4ad} shows a slight improvement but still suffers from false positives. \textit{Our method, however, achieves the lowest number of false positives and accurately localizes the small defects.} Future research should focus on enhancing robustness to background impurities and reducing false positives.

\noindent\textbf{The Effectiveness of Different Architecture with MVAS.} To validate the effectiveness of the proposed MVAS algorithm, we integrate it into different frameworks of various methods. We select three well-known methods in the field of industrial anomaly detection. Specifically, UniAD~\cite{uniad} resizes features of different scales to the smallest feature map in the intermediate layer, concatenates them, and then applies the MVAS algorithm. ViTAD~\cite{vitad} concatenates features extracted from different stages of the pre-trained ViT and then uses the MVAS algorithm. Under the RD~\cite{rd4ad} framework, features of different scales are extracted using a pre-trained ResNet-34, and the MVAS algorithm is applied to these multi-scale features. The results in Tab.~\ref{abl:archite} demonstrate that with a slight increase in parameters, FLOPs, and training memory, the model's performance shows a significant improvements in average metrics at the sample, image, and pixel levels.

\section{Conclusion}
The paper introduces the MVAD approach, the first to solve the challenging task of multi-view anomaly detection. To address the fusion and learning of multi-view features, we propose the MVAS algorithm. Specifically, the feature maps are divided into neighbourhood attention windows, and then the semantic correlated matrix between each window within single-view and all windows across multi-views is calculated. Cross-attention operations are conducted between each window in a single-view and the top-$k$ most correlated windows in the multi-view context. The MVAS can be minimized to $O((hw)^\frac{4}{3})$ computational complexity with proper window size. The entire MVAD employs an encoder-decoder architecture, incorporating MVAS blocks with varying dimensions at each feature scale and an FPN-like architecture for fusion. Extensive experiments on the Real-IAD dataset demonstrate the effectiveness and efficiency of our approach in achieving SoTA performance.

\noindent\textbf{Broader Impact.} 
We conducted a systematic investigation into the task of visual industrial multi-view anomaly detection, providing a comprehensive definition for multi-view tasks and proposing an efficient and effective attention-based methodology, aiming to contribute to the future development of the community. The diffusion model demonstrates robust generative and learning capabilities, and we intend to explore its application in multi-view anomaly detection tasks in future research.

\section*{Acknowledgments}
This work was supported by Jianbing Lingyan Foundation of Zhejiang Province, P.R. China (Grant No. 2023C01022) and Major Project of Science and Technology of Yunnan Province, China under Grant 202402AD080001.

\bibliographystyle{IEEEtran}
\bibliography{IEEEfull}

\begin{IEEEbiography}[{\includegraphics[width=1in,height=1.25in,clip,keepaspectratio]{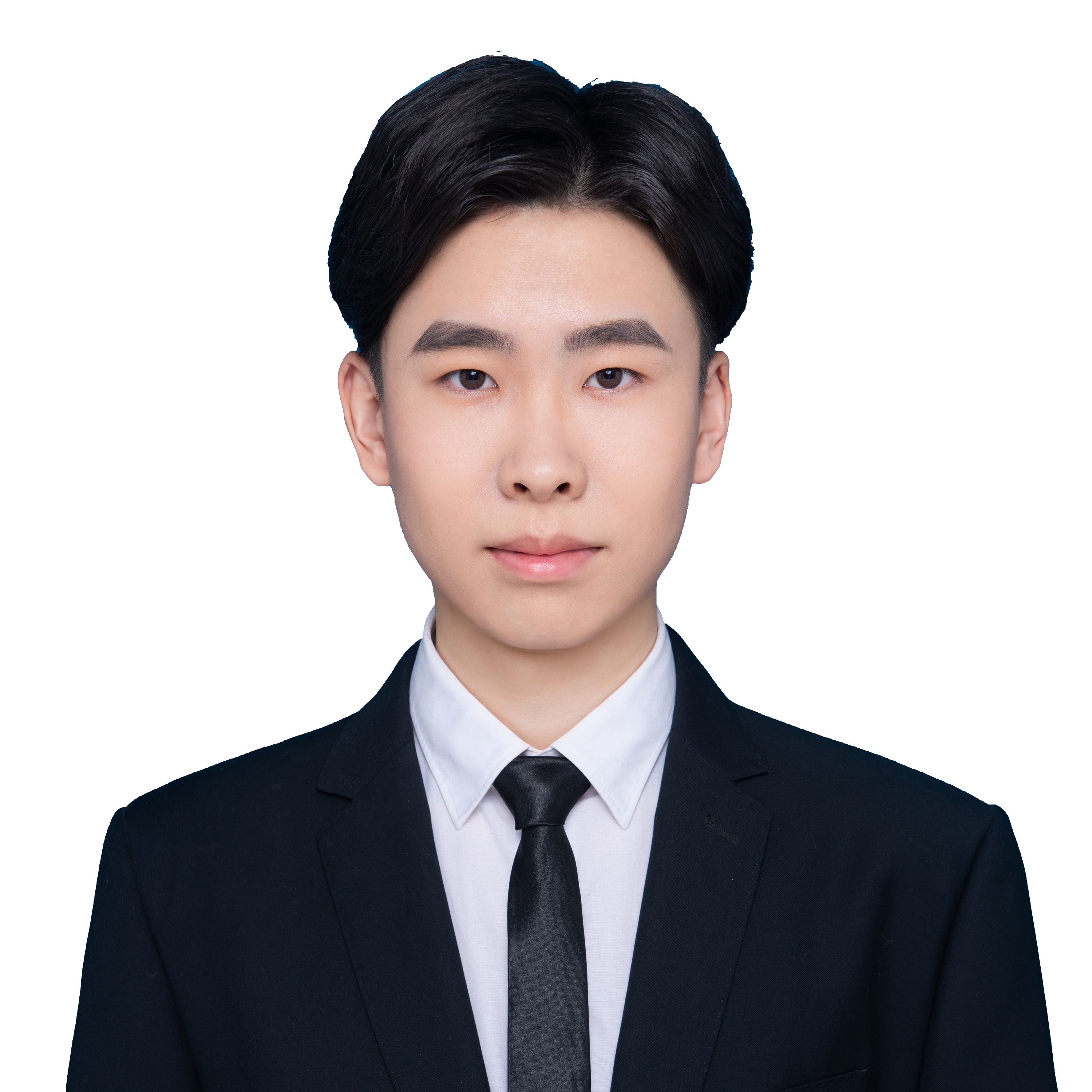}}]{Haoyang He} received the B.S. degree from Southwest Jiaotong University, Chengdu, China, in 2022. He is currently pursuing the Ph.D. degree in control science and engineering with the Institute of Cyber-Systems and Control, Zhejiang University, Hangzhou, China. His current research interests include anomaly detection, neural architecture design, and AIGC. 
\end{IEEEbiography}

\begin{IEEEbiography}[{\includegraphics[width=1in,height=1.25in,clip,keepaspectratio]{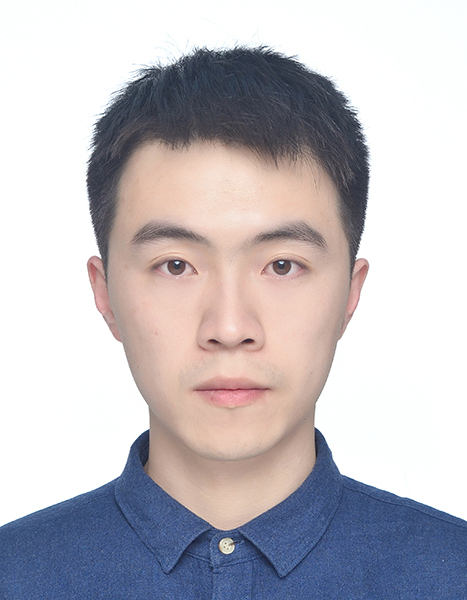}}]{Jiangning Zhang} received the B.S. degree in Electronic Information School from Wuhan University, Wuhan, China, in 2017, and the Ph.D. degree in College of Control Science and Engineering from Zhejiang University, Hangzhou, China, in 2022. He is currently a Research Scientist at Youtu Lab, Tencent, Shanghai, China. His research interests include artificial intelligence generated content and deep learning.
\end{IEEEbiography}

\begin{IEEEbiography}[{\includegraphics[width=1in,height=1.25in,clip,keepaspectratio]{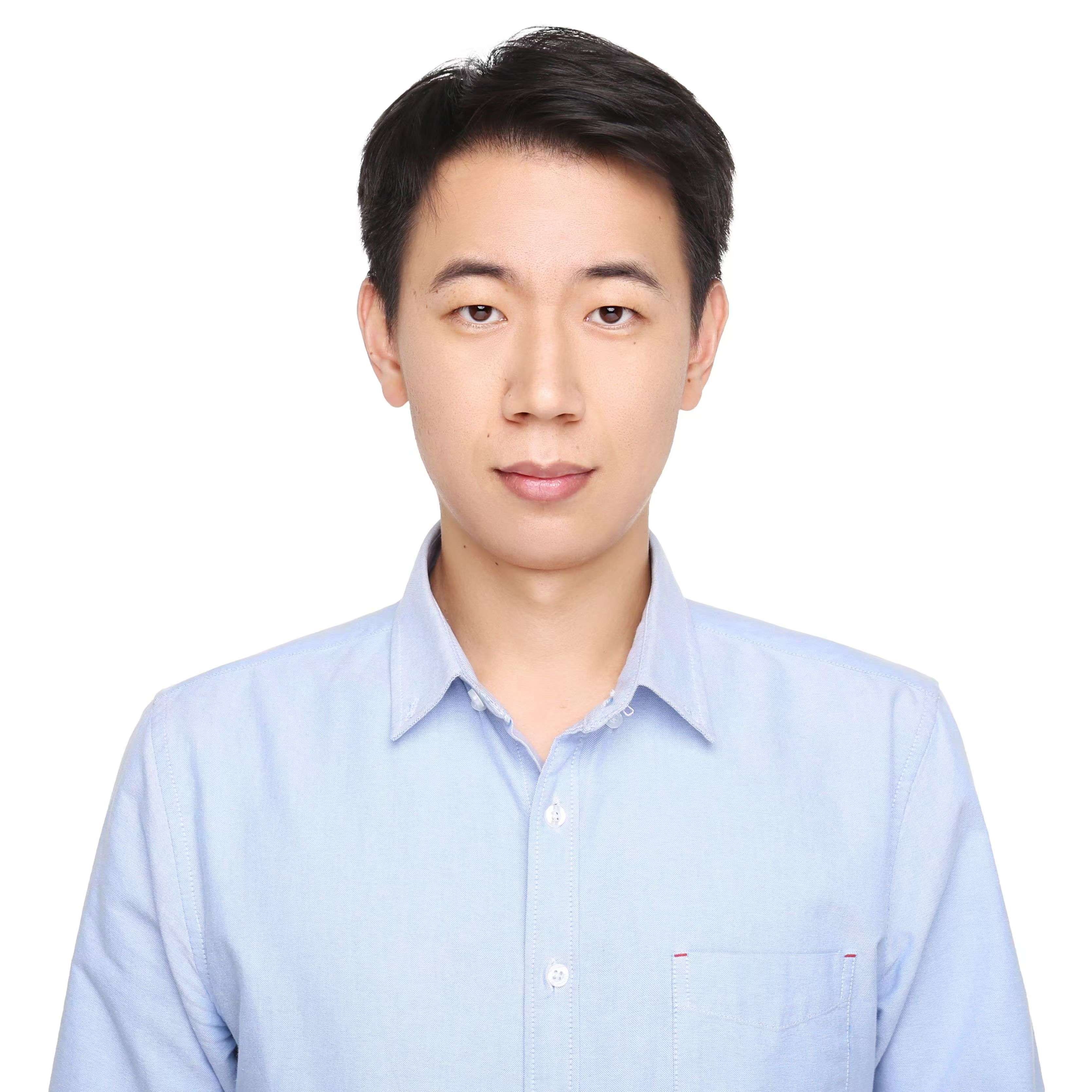}}]{Guanzhong Tian} (Member, IEEE) received the B.S. degree from Harbin Institute of Technology, Harbin, China, in 2010, and the Ph.D. degree in  from Zhejiang University, Hangzhou, China, in 2021. He is currently a Research associate with Ningbo Innovation Center, Zhejiang University. His research interests include computer vision, model compression, embodied AI.
\end{IEEEbiography}

\begin{IEEEbiography}[{\includegraphics[width=1in,height=1.25in,clip,keepaspectratio]{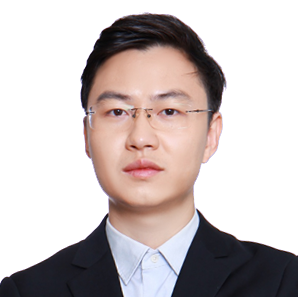}}]{Chengjie Wang} received the B.S. degree in computer science from Shanghai Jiao Tong University, China, in 2011, and double M.S. degrees in computer science from Shanghai Jiao Tong University, China and Weseda University, Japan, in 2014. He is currently the Research Director of Tencent YouTu Lab and pursuing PhD degree at Shanghai Jiao Tong University . His research interests include computer vision and machine learning. He has published more than 100 papers on major Computer Vision and Artificial Intelligence Conferences such as CVPR, ICCV, ECCV, AAAI, IJCAI and NeurIPS, and holds over 100 patents in these areas. 
\end{IEEEbiography}

\begin{IEEEbiography}[{\includegraphics[width=1in,height=1.25in,clip,keepaspectratio]{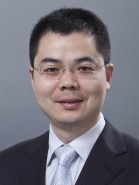}}]{Lei Xie} received a B.S. degree in 2000 and a Ph.D. in 2005 from Zhejiang University, P.R. China. Between 2005 and 2006, he was a postdoctoral researcher at Berlin University of Technology, an Assistant Professor between 2005 and 2008 and is currently a Professor at the Department of Control Science and Engineering, Zhejiang University. To date, his research activities culminated in over 30 articles that are published in internationally renowned journals and conferences, 3 book chapters and a book in the area of applied multivariate statistics and modeling. His research interests focus on the interdisciplinary area of statistics and system control theory.
\end{IEEEbiography}

\end{document}